# Comparing Haar-Hilbert and Log-Gabor Based Iris Encoders on Bath Iris Image Database

N. Popescu-Bodorin, *Member*, IEEE
Dept. of Math. & Computer Science, 'Spiru Haret' University
Bucharest, România
bodorin@ieee.org

V.E. Balas, *Senior Member*, IEEE
Faculty of Engineering, 'Aurel Vlaicu' University
Arad, România
balas@drbalas.ro

*Abstract*- **This papers introduces a new family of iris encoders which use 2-dimensional Haar Wavelet Transform for noise attenuation, and Hilbert Transform to encode the iris texture. In order to prove the usefulness of the newly proposed iris encoding approach, the recognition results obtained by using these new encoders are compared to those obtained using the classical Log-Gabor iris encoder. Twelve tests involving single/multi-enrollment and conducted on Bath Iris Image Database are presented here. One of these tests achieves an Equal Error Rate comparable to the lowest value reported so far for this database. New Matlab tools for iris image processing are also released together with this paper: a second version of the Circular Fuzzy Iris Segmentator (CFIS2), a fast Log-Gabor encoder and two Haar-Hilbert based encoders.**

I. INTRODUCTION

Nowadays, it could appear the temptation to believe that the iris recognition is a closed domain, but it shouldn't happen this way because still there are unanswered questions in this domain: what role should play the concept of *'fragile bits'* [1] in iris recognition? How many useful species of *'fragile bits'* exists? Are the binary identities really *stationary* over time? How far goes the *template aging* [2] phenomenon? Is it really necessary to take into account of *fail-to-enroll* (FTE) and *fail-to-acquire* (FTA) rates [3] when comparing matching algorithms? Is the iris recognition really a *recognition* procedure or is just *verification*? Let us take the latter question for example: if two objects have (nearly) the same shadow in a space of binary matrices, should we tell that they *match each other* in this particular sense, or should we call them (nearly) *identical objects*? Since the list of questions does not end here, we find appropriate to continue our effort in releasing new iris recognition tools and making iris recognition domain more accessible than it is today, hoping that new and interesting answers and questions will arise soon within the research community.

### A. State of the art

There are many important contributions that we would like to mention here. Still, given the subject of our paper, we will rely mostly on works related to the *University of Bath Iris Image Database* (UBIID). Two of the reference works on UBIID was undertaken in [4] and [5]. The values estimated in these papers for the *Equal Error Rate* (EER) continue to be the lowest announced so far for this database (1.2E-4 and 7.5E-5, respectively). Both papers insist especially on the accuracy of the segmentation. On the contrary, our approach assumes an inherent imprecision of the segmentation: the iris is considered to be a pupil-concentric circular ring. The question is if this rough approximation of the iris still can guarantee good recognition results.

On the other hand, one of the most comprehensive study ever undertaken in iris recognition is the so called 'IREX I' report [3] which is mainly an independent evaluation of the main recognition algorithms available on the market of biometric solutions. Since our paper does not take into account for FTA and FTE rates, we will give a special attention to the data reported in [3] for the following algorithms: SAGEM (A1, A2), COGENT (B1, B2) and IRITECH (I1, I2) for which the FTA and FTE rates are both nil. For these algorithms, the *Generalized False Accept Rate* (GFAR) and the *Generalized False Reject Rate* (GFRR) equal the *False Match Rate* (FMR) and *False Non-Match Rate* (FNMR), respectively (see Section 7.3.3 in [3]). Consequently, the *Receiver Operating Characteristic* (ROC) data reported for these algorithms are indeed suitable for comparative testing of algorithms despite the fact that this report sometimes said otherwise (see Figure 9 in [3]), and despite the fact that, sometimes, GFAR and GFRR are used as ROC coordinates instead of FMR and FNMR. Alternatively, some authors prefer to rename FMR and FNMR as *False Accept Rate* (FAR) and *False Reject Rate* (FRR), while others use the first set of names when talk about iris templates and the second set when talk about persons. We belong in the former category: here FAR and FRR have the same meaning as FMR and FNMR, respectively. A very good image of the state of the art algorithms mentioned above is given by the data presented in Figure 9 and Figure 11 from [3]. It can be seen that the state of the art algorithms of these days negotiate *system accuracy* and *user comfort* in terms of FAR (in range of $10^{-1}$-$10^{-6}$) and FRR (in range of 0-5%). The question is if the methods proposed here can lead to similar results.

Regarding the performance evaluation methodology, we rely on Daugman's works [6] and [7], but we also take into account of the remark formulated in Section 6.3 from [3] that we fully agree with: using decidability index as a measure of the separation between genuine and imposter score distributions "is appropriate when the distributions are Normal but is less relevant otherwise because it does not capture the functional form at the overlap" of their tails. Consequently, in order to preserve as much as possible the normality of the imposter score distribution, the (unmodified)

Hamming distance is preferred here to express the similarity between any two iris codes. Doing so is consistent with the hypothesis formulated by Daugman that the matching between different irides happens only *by chance* [7].

## II. PROPOSED METHODS

### A. CFIS2 Segmentation Procedure

The segmentation procedure used and tested in this paper, abbreviated CFIS2, is a faster variant of Circular Fuzzy Iris Segmentation (CFIS, [8], [9]). It allows to extract a fuzzy limbic boundary [8] ten times faster than finding the pupil (i.e. in 5-6 milliseconds when running as a Matlab script on a P-IV Prescott processor at 2.8GHz). Speed optimization was made possible by using the test infrastructure described in [9] while searching for a heuristic procedure to reduce the search space when seeking for the limbic boundary. The fully functional CFIS2 Matlab function is available for download together with the second version of the Processing Toolbox for the University of Bath Iris Image Database (PT-UBIID-v.02, [10]). In this moment, limbic boundary detector is the most optimized part of our segmentation procedure and future improvement of the other components is just a matter of time (future collaborative works are welcomed).

Briefly, the limbic boundary detector within CFIS2 algorithm is presented in Fig.1.a, and formulated as follows:

**Fast Limbic Boundary Detector** (N. Popescu-Bodorin)
**Input:** unwrapped iris region and pupil radius;
1. Crop two subimages of unwrapped iris region, each of them covering the range of 22.5°-45° under horizontal diameter, on the left and right sides, respectively. For each of them:
1.a. Crop the subimage by eliminating the pupil region.
1.b. Compute the vector **V** containing the means of all columns within the image resulting from step (2); Find that local maximum of **V** which is the closest to the pupil. Crop the image again at the index of that maximum point.
1.c. Compute 2-means binarization of the subimage obtained in the previous step. A vertical separator of the two regions within the binarized image gives the index $i_k$ (k=1 or k=2) of the fuzzy limbic boundary.
2. Take the index **i** of the fuzzy limbic boundary as being the minimum between $i_1$ and $i_2$.
**Output:** the index **i** of the fuzzy limbic boundary.

Despite its simplicity, the above algorithm is very effective. Indeed, it fails to find the limbic boundary for only two cases from a total of 1000 images contained in UBIID. Hence, its failure rate on the UBIID is 2‰. The pupil finder [9] also fails for one image within the database. For these reasons, the failure rate of CFIS2 is 3‰. During the tests undertaken to verify the results of our limbic boundary detector, we saw that the left and right eyes of the subject 17 are, in fact, interchanged.

### B. Log-Gabor Encoder

The *Log-Gabor Encoder* (LGE) [10] used here is a faster and simpler variant of the encoder designed originally by Libor Masek [11]. Our version is a single-scale one-dimensional Log-Gabor filter. Each line of the unwrapped normalized iris segment situated in the angular direction is

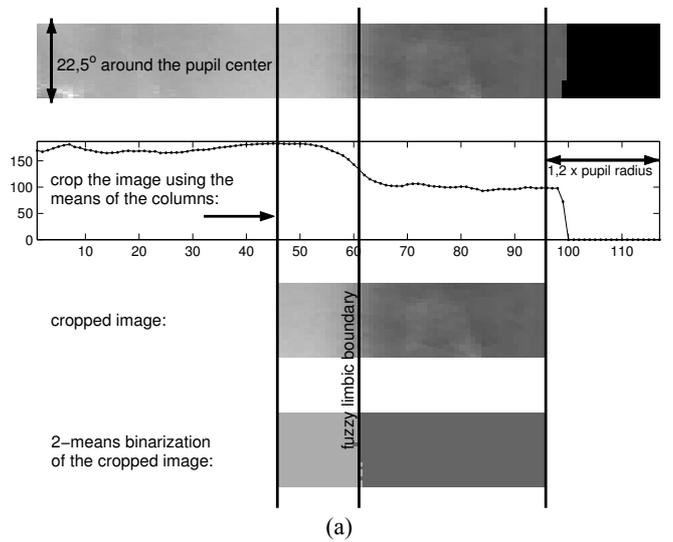

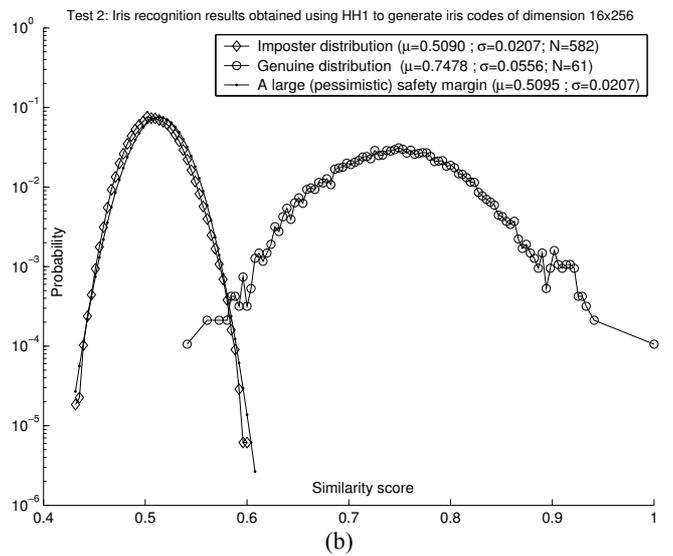

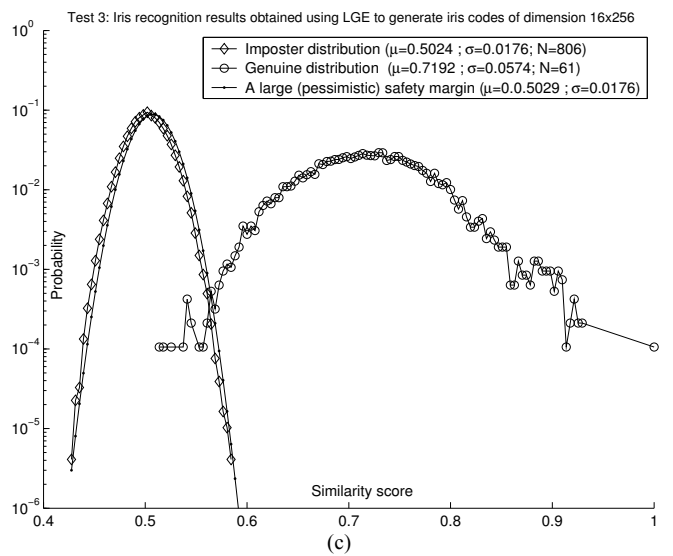

Fig. 1. The results of two recognition tests which use binary iris codes of dimension 16x256 (4Kb) generated with HH1 (b) and LGE (c), respectively. The segmentation procedure used in all tests is CFIS2 (a).

convolved with 1-D Log-Gabor Wavelet:

$$G(f) = \exp[-0.5 \cdot \log^2(f/f_0) / \log^2(\sigma/f_0)] \quad (1)$$

where $f_0$ and $\sigma$ represent the center frequency and the bandwidth of the filter. Prior to the Log-Gabor filter we don't make any enhancement of the iris texture. The encoder is also available online [10].

### C. Haar-Hilbert based encoders

Haar-Hilbert based iris encoders presented here use single-level 2-dimensional Discrete Haar Wavelet Transform for noise attenuation and Hilbert Transform to encode the phase content within the iris texture. Our experimental work showed that if we practice a single-level wavelet decomposition of the unwrapped iris segments and if Hilbert Transform [9] is used to generate binary codes for each frame of the decomposition, then the detail frames representing the same iris are as similar to each other as those coming from different irides. In other words, in the context of iris recognition, the information stored in the detail frames of the wavelet decomposition is mainly noise. On the contrary, the Hamming distances between the iris codes generated using Hilbert Transform for the approximation frames coming from the images of the same iris are smaller than the Hamming distances between the iris codes generated directly from the initial iris segments. Unfortunately, this is not the case when LGE is used to generate binary iris codes. Briefly, the first Haar-Hilbert encoder used here can de stated as follows:

**HH1 Encoder** (N. Popescu-Bodorin)
**Input:** unwrapped normalized iris segment **NI** with lines representing the angular direction, and the dimension **s** of the Hilbert filter;
1. Retrieve approximation frame **AF** from the single-level 2-dimensional Haar wavelet decomposition of **NI**.
2. Reshape **AF** as a matrix **R** with s lines such that the lines of **AF** to fill the columns of **R**.
3. Consider the matrix **H** whose columns are computed by applying the Hilbert Transform column-wise in **R**.
4. Compute the column-wise analytic signal **AS** = **R** + **H**.
5. Generate the binary iris code **IC** as the logical index of all those components within **AS** that have positive imaginary part:
   IC = logical (0 < imag(**AS**)).
**Output:** the binary iris code **IC**.

The second Haar-Hilbert encoder presented here uses two Hilbert filters of different sizes:

**HH2 Encoder** (N. Popescu-Bodorin)
**Input:** unwrapped normalized iris segment **NI** with lines in the angular direction, and the even dimension **s** of the Hilbert filter;
1. Retrieve approximation frame **AF** from the single-level 2-dimensional Haar wavelet decomposition of **NI**.
2. Reshape **AF** as a matrix **R** with s lines such that the lines of **AF** to fill the columns of **R**.
3. Apply the Hilbert Transform column-wise in **R**. to obtain **H$_1$**
4. Consider the matrix **H$_2$** obtained by applying the step 3 for the top and bottom half parts of **R**.
5. Compute the sum **AS** = 2**R** + **H$_1$** + **H$_2$**.
6. Generate the binary iris code **IC** as the logical index:
   IC = logical (0 < imag(**AS**)).
**Output:** the binary iris code **IC**.

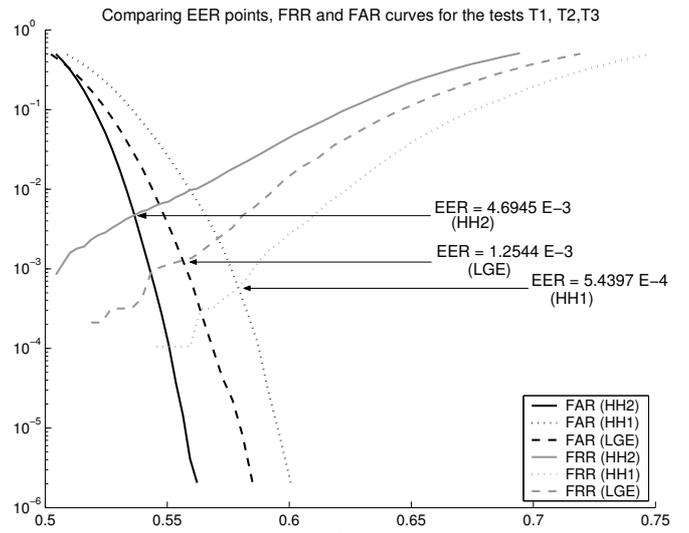

Fig. 2. EER points for the tests T1 (HH2), T2 (HH1) and T3 (LGE).

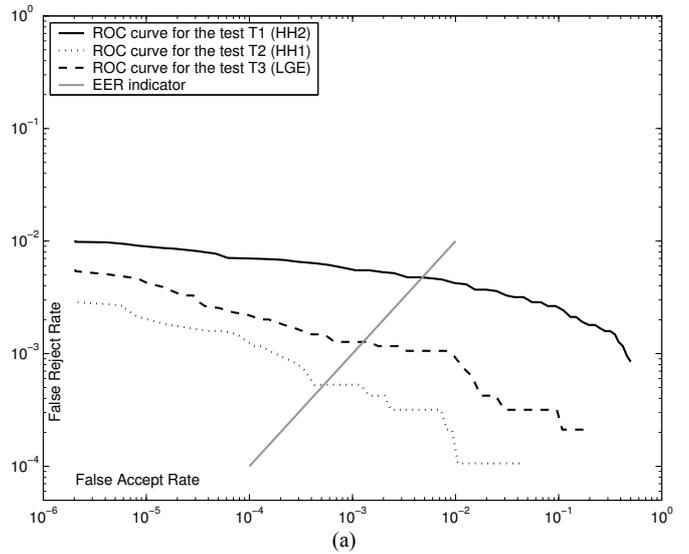

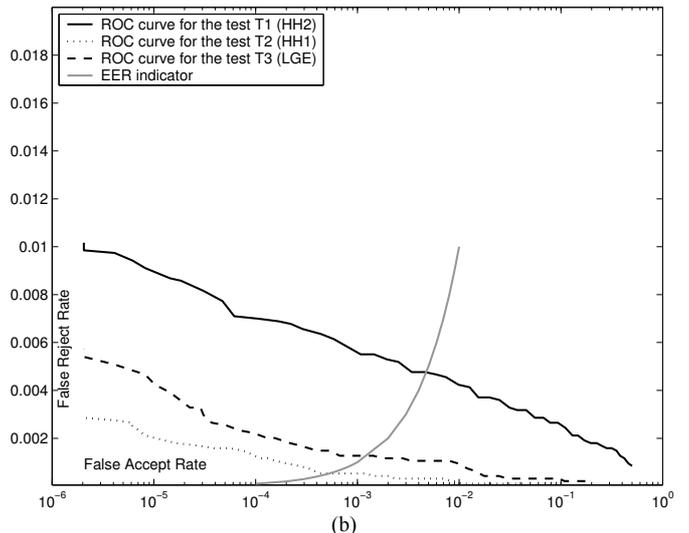

Fig. 3. ROC curves for the tests T1 (HH2), T2 (HH1) and T3 (LGE): log-log plot with better resolution near EER, on the right side of EER (a), and semilogx plot with better resolution on the left side of EER points (b). All of these tests use binary iris codes of size 16x256 (4 Kb).

## III. PERFORMANCE METRICS

The quality of a biometric system is given by the 'distance' between genuine and imposter score distribution. Intuitively, we know that the separation between the distributions of scores is better when the distance between their means increases while their variance decreases. This is usually expressed as decidability index **d'** [12] and Fisher ratio **r** [13]:

$$d' = |\mu_i - \mu_g| / [0.5 \cdot (v_i + v_g)]^{1/2} \quad (2)$$
$$r = (\mu_i - \mu_g)^2 / (v_i + v_g) \quad (3)$$
$$d' = (2 \cdot r)^{1/2} \quad (4)$$

From a practical point of view, it is also important to know Equal Error Rate (EER - the common value of FAR and FRR at the point where they equal each other), FRR values corresponding to certain FAR values (such as: $10^{-1}$, $10^{-2}$, $10^{-3}$, $10^{-4}$, $10^{-5}$ and $10^{-6}$), and FAR values corresponding to a set of given FRR values (in range of 1-5%). Ideally, a perfect recognition system should have a narrow imposter score distribution and a low EER point near which the FRR slope to be as weak as possible in a vicinity as wide as possible.

Another important separation measure is the value of FAR at the minimum genuine score (FAR at the first nonzero FRR) and FRR at the last nonzero FAR (FRR at the maximum imposter score). Al of these performance metrics are used here to evaluate the methods discussed.

Pessimistic Odds of False Accept (POFA), OFA and Odds of False Reject (OFR) [9] are computed as follows:

$$POFA(t) = \int_t^1 PI_{pdf} d\tau,$$

$$OFA(t) = \int_t^1 I_{pdf} d\tau, \quad OFR(t) = \int_0^t G_{pdf} d\tau$$

where $I_{pdf}$ and $G_{pdf}$ are the measured probability density functions of the imposter and genuine distribution respectively, **t** is the threshold and $PI_{pdf}$ is a theoretical pessimistic imposter probability density function chosen such as to be an upper bound envelope for the measured imposter score distribution on its right side, starting with $10^{-2}$ and continuing downwards. For all tests presented here, these conditions are fulfilled by a normal distribution having the same standard deviation as the measured imposter distribution and a mean shifted to the right with 5E-3.

Regarding the manner of displaying the score distributions, we think that semilogy plots (Fig.1.b, Fig.1.c, and Fig.4.a - Fig.4.c) show more clearly what happens in the region where their tails are or could be overlapped. We also use log-log and semilogx ROC plots (Fig.3.a, Fig.3.b) in order to ensure a better resolution near EER point, both on the left and right sides.

## IV. THE TESTS

The tests undertaken here are organized in two series: 6 single-enrollment tests (T1-T6) and 6 multi-enrollment tests (T7-T12). Each series contains 3 tests (T1-T3, T7-T9) that use bigger iris codes of size 16x256 (4Kb), and 3 tests (T4-T6, T10-T12) that use smaller iris codes (1Kb). Their results

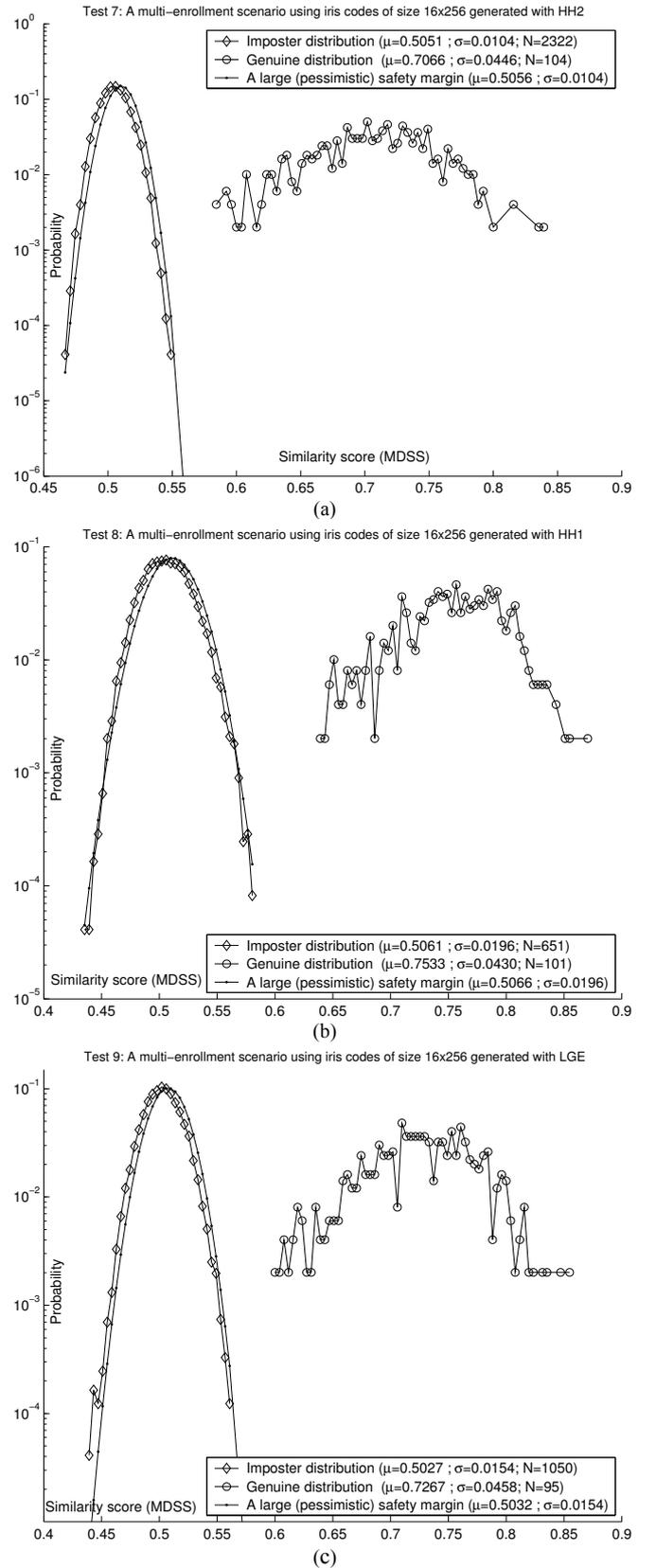

Fig. 4. Iris recognition results for the tests T7 (a), T8 (b) and T9 (c). Iris codes of size 16x256 are generated used HH2, HH1 and LGE, respectively. All candidate templates are compared to all enrolled identities.

TABLE 1: SIX SINGLE-ENROLLMENT IRIS RECOGNITION TESTS ON BATH IRIS IMAGE DATABASE

| TESTS: | T1 | T2 | T3 | T4 | T5 | T6 |
|---|---|---|---|---|---|---|
| **System parameters:** | | | | | | |
| Iris code size | 16x256 (4Kb) | 16x256 (4Kb) | 16x256 (4Kb) | 16x64 (1Kb) | 8x128 (1Kb) | 8x128 (1Kb) |
| Encoder/Segmentator | HH2 / CFIS2 | HH1 / CFIS2 | LGE / CFIS2 | HH2 / CFIS2 | HH1 / CFIS2 | LGE / CFIS2 |
| Hilbert filter size | 16; 8 | 16 | Not applicable | 16; 8 | 8 | Not applicable |
| **Inter-class score distribution:** | | | | | | |
| Mean / Standard deviation | 0.5044 / 0.0122 | 0.5090 / 0.0207 | 0.5024 / 0.0176 | 0.5067 / 0.0226 | 0.5057 / 0.0212 | 0.5023 / 0.0223 |
| Degrees-of-freedom | 1'681 | 582 | 806 | 491 | 555 | 502 |
| **Intra-class score distribution:** | | | | | | |
| Mean / Standard deviation | 0.6945 / 0.0585 | 0.7478 / 0.0556 | 0.7192 / 0.0574 | 0.7368 / 0.0552 | 0.7303 / 0.0549 | 0.7213 / 0.0596 |
| Degrees-of-freedom | 62 | 60 | 61 | 64 | 65 | 57 |
| **Evaluation criteria:** | | | | | | |
| Decidability index / Fisher's ratio | 4.5 / 10.1248 | 5.689 / 16.1826 | 5.1111 / 13.0617 | 5.4523 / 14.8636 | 5.3947 / 14.5511 | 4.8657 / 11.8377 |
| EER | 4.6945 E-3 | **5.4397E-4** | 1.2544E-3 | 1.504E-3 | 2.0747E-3 | 2.2725E-3 |
| MIS (Maximum Imposter Score) | 0.5623 | 0.6006 | 0.5850 | 0.6162 | 0.6064 | 0.6064 |
| FAR(MIS) / FRR(MIS) | **2.053E-6 / 1.02E-2** | **2.053E-6 / 2.9E-3** | **2.053E-6 / 5.7E-3** | 2.053E-6 / 1.41E-2 | 2.053E-6 / 1.25E-2 | 2.053E-6 / 2.5E-2 |
| Storage efficiency | 41.04 % | 14.21% | 19.68% | 47.95% | 54.20% | 49.02% |
| **FUNCTIONING REGIMES:** | | | | | | |
| **near a FRR of 0.01:** | | | | | | |
| threshold / FRR(threshold) | 0.56176 / 9.9545E-3 | 0.62359 / 9.8486E-3 | 0.59396 / 9.5309E-3 | 0.60921 / 9.7427E-3 | 0.60045 / 9.6368E-3 | 0.58845 / 9.8486E-3 |
| FAR(threshold) / POFA(threshold) | **0** / 8.7648E-6 | **0** / 6.1826E-8 | **0** / 4.4116E-7 | 6.159E-6 / 7.857E-6 | 4.106E-6 / 1.172E-5 | 5.338E-5 / 1.392E-4 |
| **near a FRR of 0.02:** | | | | | | |
| threshold / FRR(threshold) | 0.57826 / 0.019168 | 0.63559 / 0.019168 | 0.60596 / 0.019379 | 0.62033 / 0.019591 | 0.61745 / 0.019591 | 0.60245 / 0.019803 |
| FAR(threshold) / POFA(threshold) | **0** / 8.1463E-9 | **0** / 2.2201E-9 | **0** / 1.0869E-8 | **0** / 3.9304E-7 | **0** / 2.4487E-7 | 4.106E-6 / 1.014E-5 |
| **near a FRR of 0.03:** | | | | | | |
| threshold / FRR(threshold) | 0.58926 / 0.029016 | 0.64459 / 0.029863 | 0.61396 / 0.028381 | 0.63033 / 0.029122 | 0.62545 / 0.028804 | 0.61045 / 0.029228 |
| FAR(threshold) / POFA(threshold) | **0** / 2.8875E-11 | **0** / 1.478E-10 | **0** / 7.1623E-10 | **0** / 4.7091E-8 | **0** / 3.1999E-8 | **0** / 1.9181E-6 |
| **near a FRR of 0.04:** | | | | | | |
| threshold / FRR(threshold) | 0.59726 / 0.039606 | 0.64959 / 0.037912 | 0.61996 / 0.038653 | 0.63733 / 0.038971 | 0.63445 / 0.039394 | 0.61745 / 0.038865 |
| FAR(threshold) / POFA(threshold) | **0** / 2.891E-13 | **0** / 3.0295e-011 | **0** / 8.1619e-011 | **0** / 8.126E-9 | **0** / 2.7493E-9 | **0** / 4.0353E-7 |
| **near a FAR of 1E-3:** | | | | | | |
| threshold / FRR(threshold) | 0.54276 / 5.5067E-3 | 0.57659 / 5.2949E-4 | 0.55796 / 1.2708E-3 | 0.57921 / 2.0121E-3 | 0.57245 / 2.7534E-3 | 0.57145 / 3.8123E-3 |
| FAR(threshold) / POFA(threshold) | 1.036E-3 / 3.108E-3 | 9.711E-4 / 1.261E-3 | 9.931E-4 / 2.038 E-3 | 9.403E-4 / 1.399E-3 | 9.916E-4 / 1.807E-3 | 9.690E-4 / 2.033E-3 |
| **near a FAR of 1E-4:** | | | | | | |
| threshold / FRR(threshold) | 0.55126 / 7.095E-3 | 0.58759 / 1.1649E-3 | 0.56846 / 2.1180E-3 | 0.59521 / 4.5536E-3 | 0.58595 / 5.8244E-3 | 0.58445 / 8.0483E-3 |
| FAR(threshold) / POFA(threshold) | 9.23E-5 / 2.98E-4 | 1.026E-4 / 1.916E-4 | 1.047E-4 / 2.618E-4 | 1.026E-4 / 1.086E-4 | 9.649E-5 / 1.955E-4 | 8.623E-5 / 2.748E-4 |
| **near a FAR of 1E-5:** | | | | | | |
| threshold / FRR(threshold) | 0.55676 / 8.6837E-3 | 0.59509 / 2.0121E-3 | 0.57946 / 4.130E-3 | 0.60671 / 8.7896E-3 | 0.59595 / 8.4719E-3 | 0.59695 / 1.5779E-2 |
| FAR(threshold) / POFA(threshold) | 1.026E-5 / 5.133E-5 | 1.026E-5 / 4.554E-5 | 1.026E-5 / 2.129E-5 | 1.026E-5 / 1.290E-5 | 1.026E-5 / 2.944E-5 | 8.212E-6 / 2.968E-5 |

TABLE 2: SIX MULTI-ENROLLMENT IRIS RECOGNITION TESTS ON BATH IRIS IMAGE DATABASE

| TESTS: | T7 | T8 | T9 | T10 | T11 | T12 |
|---|---|---|---|---|---|---|
| **System parameters:** | | | | | | |
| Iris code size | 16x256 (4Kb) | 16x256 (4Kb) | 16x256 (4Kb) | 16x64 (1Kb) | 8x128 (1Kb) | 8x128 (1Kb) |
| Encoder/Segmentator | HH2 / CFIS2 | HH1 / CFIS2 | LGE / CFIS2 | HH2 / CFIS2 | HH1 / CFIS2 | LGE / CFIS2 |
| Hilbert filter size | 16; 8 | 16 | Not applicable | 16; 8 | 8 | Not applicable |
| **Inter-class score distribution:** | | | | | | |
| Mean / Standard deviation | 0.5051 / 0.0104 | 0.5061 / 0.0196 | 0.5027 / 0.0154 | 0.5071 / 0.0197 | 0.5030 / 0.0181 | 0.5051 / 0.0185 |
| Degrees-of-freedom | 2'322 | 651 | 1'050 | 642 | 765 | 734 |
| **Intra-class score distribution:** | | | | | | |
| Mean / Standard deviation | 0.7066 / 0.0446 | 0.7533 / 0.0430 | 0.7267 / 0.0458 | 0.7416 / 0.0429 | 0.7360 / 0.0422 | 0.7283 / 0.0472 |
| Degrees-of-freedom | 104 | 101 | 95 | 104 | 110 | 89 |
| **Evaluation criteria:** | | | | | | |
| Decidability / Fisher's ratio / EER | 6.22 / 19.35 / **0** | 7.40 / 27.38 / **0** | 6.55 / 21.50 / **0** | 7.014 / 24.60 / **0** | 7.05 / 24.88 / **0** | 6.23 / 19.42 / **0** |
| MIS / mGS (minimum genuine score) | 0.5495 / 0.5843 | 0.5788 / 0.6407 | 0.56217 / 0.60097 | 0.59117 / 0.62222 | 0.57708 / 0.61451 | 0.57330 / 0.59256 |
| POFA (mGS) | **4.412E-13** | **1.9101e-011** | **7.5929E-10** | **1.1993E-8** | **1.9068E-9** | **3.9024E-6** |
| Storage efficiency | 56.69 % | 15.89 % | 25.63 % | 62.69 % | 74.71 % | 71.68 % |
| **FUNCTIONING REGIMES:** | | | | | | |
| **near a FRR of 0.01:** | | | | | | |
| threshold / FRR(threshold) | 0.59354 / 6.0362E-3 | 0.64881 / 6.0362E-3 | 0.61017 / 8.0483E-3 | 0.62817 / 6.0362E-3 | 0.62508 / 8.0483E-3 | 0.6063 / 8.0483E-3 |
| FAR(threshold) / POFA(threshold) | **0** / **4.4409E-16** | **0** / 1.0804E-12 | **0** / 1.5808E-11 | **0** / 2.0259E-9 | **0** / 4.6686E-11 | **0** / 9.1918E-8 |
| **near a FRR of 0.02:** | | | | | | |
| threshold / FRR(threshold) | 0.60554 / 0.018109 | 0.65181 / 0.016097 | 0.61917 / 0.018109 | 0.64217 / 0.018109 | 0.63308 / 0.018109 | 0.6163 / 0.018109 |
| FAR(threshold) / POFA(threshold) | **0** / **0** | **0** / 3.5694E-13 | **0** / 2.558E-13 | **0** / 2.1728E-11 | **0** / 2.2587E-12 | **0** / 4.2851E-9 |
| **near a FRR of 0.03:** | | | | | | |
| threshold / FRR(threshold) | 0.61554 / 0.028169 | 0.66081 / 0.028169 | 0.62617 / 0.028169 | 0.65017 / 0.028169 | 0.64408 / 0.028169 | 0.6213 / 0.026157 |
| FAR(threshold) / POFA(threshold) | **0** / **0** | **0** / 1.1213e-014 | **0** / 8.2157E-15 | **0** / 1.3043E-12 | **0** / 2.5646E-14 | **0** / 8.3117E-10 |

are presented in Table I and Table II, but also in Fig.1.a - Fig.4.c. Each single/multi-enrollment test counts 487'063/ 24'353 unique imposter comparisons and 9'443/497 unique genuine comparisons, respectively. In addition to the notations already discussed so far, the following acronyms appear in the tables: MIS - maximum imposter score, and mGS - minimum genuine scores. When MIS is greater than mGS the measured score distributions overlap each other. Hence another performance metric is the difference mGS-MIS. It can be seen in Fig.1.b, Fig1.c and Table I, that the measured imposter and genuine score distributions overlap each other for all single-enrollment tests. A different situation is observed in the multi-enrollment tests (Table II, tests T7-T12). For all of them the measured imposter and genuine score distributions do not overlap each other. In these multi-enrollment tests, the motivation behind the better separation of the measured imposter and genuine distributions is the fact that each identity is defined as a collection of 10 binary templates, hence it implicitly includes in its definition an inherent variability of this 'matching by chance' game. Consequently the distance between a candidate template and an enrolled identity is given not only by the Hamming distances to each individual templates enrolled under that identity, but also by the mean and the standard deviation of the set formed with these Hamming distances. In these conditions the Mean-Deviation Similarity Score (MDSS) between a candidate template (iris code) and an enrolled identity is defined as:

$$MDSS(C,EI) = mean(HSS) + std(HSS) - s/2 \quad (5)$$

where **s** is the standard deviation of the imposter distribution measured in the corresponding single-enrollment test, and HSS is the set of all Hamming similarity scores (HS) between the candidate template C and each template IC stored under the enrolled identity EI:

$$EI = \{IC_1, ..., IC_{10}\}; HSS = \{HS(C,IC_1), ..., HS(C,IC_{10})\} \quad (6)$$

## V. COMPARING THE TESTS RESULTS

The first remarkable aspect regarding the results presented in Table I is that the values measured for FAR(MIS) in all single-enrollment tests coincide perfectly (the common value of these parameters is 2.053E-6), regardless the encoder used to generate the iris codes, regardless the iris code size (4Kb, 1Kb). Hence, FAR(MIS) is a measure of all errors accumulated in the biometric system prior to the matching and prior to the binary encoding of iris texture. In other words, FAR(MIS) is a measure of how *accurate* are the operations performed before the encoding, i.e. during acquisition, pre-processing (if any), segmentation, unwrapping and normalization. Surprisingly, the fuzzy circular pupil-concentric segmentation used here (CFIS2) is not just simple, fast and effective, but also better than one might hope at first sight. The difference between the extracted circular iris rings and the actual eccentric iris segments is not so important as long as the segmentation procedure makes the same mistake equally against all eye images. Similar behaviors were observed during the single-enrollment tests for FRR(mGS) - which preserves the value of 1.0590E-4 regardless the encoder used and the size of iris codes, and during the multi-enrollment tests for both FAR(MIS) and FRR(mGS).

On the contrary, the values of EER, FRR(MIS), FAR(mGS) and POFA(FRR$^{-1}$(c)) (where c is a parameter in range of 0.01-0.05, which establishes the degree of user comfort), all of them depend both on the encoder used to generate binary iris codes and on the iris code size.

For comparison with the results reported in 'IREX I' [3] for UBIID, we point out to Fig.9.a - Fig.9.f and Fig.11 in [3]. Despite it is said there that the figures should not be used for comparing iris recognition algorithms for the reason that the figures do not take into account the FTE and FTA rates, this argument is not sound at all since, as is mentioned in the section 7.3.3 [3] and supported by the data in Table 6 [3], FTE=FTA=0 for each of the following algorithms: SAGEM (A1, A2), COGENT(B1, B2), IRITECH(I1, I2). Hence at least the figures describing the behavior of these algorithms are indeed suitable for comparisons. Still we do not insist on these aspects mainly because 'what iris recognition really is' and 'how the biometric commercial solutions compete to each other' are two very different subjects.

Among the multi-enrollment tests using codes of 4Kb (T7-T9), it can be seen that HH2 encoding (T7) leads to the best results. Still, the decidability index is greater when HH1 encoder is used (T8). Hence, the tests T7 and T8 illustrate a limitation of decidability index as performance metric. Here, HH2 encoder is designed to minimize the standard deviation of the imposter distribution while HH1 is designed to maximize decidability index. Among the multi-enrollment tests using codes of 1Kb, HH1 encoder ensures a better negotiation between FRR and POFA values. Also, regardless the iris code size, in all single/multi-enrollment tests presented here, LGE is outperformed by a HH encoder.

## VI. CONCLUSION

This paper presented state of the art iris recognition results for UBIID [14] obtained using a fuzzy circular pupil-concentric iris segmentation procedure (CFIS2) and the newly proposed Haar-Hilbert based encoders (HH1 and HH2). The comparison to the results obtained by other authors on the same database and the comparisons made here between the Haar-Hilbert and the Log-Gabor based encoders sustain this affirmation.

ACKNOWLEDGMENT

The authors would like to thank Professor Donald Monro (Dept. of Electronic and Electrical Engineering, University of Bath, UK) for granting the access to the Bath University Iris Image Database.